\documentclass[letterpaper]{article} %
\usepackage[preprint]{aaai2027}  %
\usepackage[hyphens]{url}  %
\usepackage{graphicx} %
\usepackage{natbib}  %
\usepackage{caption} %
\usepackage{booktabs}
\usepackage{colortbl}
\usepackage{amsmath,amssymb}

\usepackage{etoolbox}
\makeatletter
\patchcmd{\@maketitle}{These authors contributed equally.}{Equal contribution.}{}{%
  \PackageError{arxiv}{equalcontrib footnote patch 1/2 failed: string not found in \string\@maketitle}{}}
\patchcmd{\@maketitle}{These authors contributed equally.}{Equal contribution.}{}{%
  \PackageError{arxiv}{equalcontrib footnote patch 2/2 failed}{}}
\makeatother

\newcommand{\METHOD}{DreamTraj}          %
\newcommand{\DATASET}{\textsc{Move}}     %
\title{DreamTraj: Generating 6-DoF Object Trajectories by \\ Reading Unrendered Video Diffusion Latents}
\author{
    Tongsheng~Ding\textsuperscript{1}\equalcontrib,
    Zhen~Luo\textsuperscript{2,1}\equalcontrib,
    Yixuan~Yang\textsuperscript{1}\equalcontrib,
    Boyu~Wang\textsuperscript{1}, \\
    Luyang~Xie\textsuperscript{1},
    Jinyu~Yang\textsuperscript{3}\corresponding,
    Feng~Zheng\textsuperscript{1,4}
}
\affiliations{
    \textsuperscript{1}Southern University of Science and Technology \quad
    \textsuperscript{2}Shanghai Innovation Institute \\
    \textsuperscript{3}Harbin Institute of Technology, Shenzhen \quad
    \textsuperscript{4}SpatialtemporalAI \\
    12532585@mail.sustech.edu.cn, luoz2024@mail.sustech.edu.cn, arnoldyang97@gmail.com
}

\newcommand{\transunit}{centimetres}
\newcommand{\mainonly}[1]{}

\begin{document}

\maketitle

\begin{abstract}
 Accurate prediction of object trajectories during manipulation 
 is essential for closing the perception–action loop.
Progress is limited on two fronts: available datasets lack fine-grained language-to-motion annotations, and existing predictors either rely on privileged inputs such as video, depth, or CAD models, or recover motion from fully generated videos through costly, error-prone perception pipelines. We close the supervision gap with the \textbf{\DATASET{}} dataset, 5,038 object-centric egocentric trajectories, each paired with a fine-grained natural-language instruction rather than a coarse verb--noun label.
We further propose \textbf{\METHOD{}}, which predicts a 6-DoF object trajectory from a single RGB image and a task instruction, requiring no video, depth, or CAD model at inference: rather than generating a video, it reads motion from the internal representations of a frozen image-to-video diffusion model at an early denoising step.
A lightweight flow-matching Reader decodes query--key attention tracks and pooled hidden states into relative 6-DoF poses.
To our knowledge, this is the first approach to directly decode object 6-DoF trajectories from intermediate video diffusion representations rather than generated pixels. \METHOD{} sets a new state of the art on both translation and rotation against forecasters that consume multi-frame or privileged inputs, and runs $4.6\times$ faster than generate-then-extract pipelines.
\end{abstract}

\begin{links}
    \link{Project page}{https://whathappen0.github.io/DreamTraj/}
\end{links}

\section{Introduction}
\label{sec:intro}

Anticipating how objects will move is a cornerstone capability for embodied intelligence. 
A robot that can predict the future trajectory of a manipulated object---before the motion actually happens---can plan grasps and placements in advance, imitate human demonstrations at the object level rather than the pixel level, and verify whether an intended action will achieve its goal. 
Recent work has repeatedly shown that object-centric SE(3) trajectories form a compact, executable interface between perception and control: once the 6-DoF motion of the target object is known, off-the-shelf controllers can carry out the manipulation~\citep{rigvid,track2act,gen2act}.
The trajectories that matter for manipulation, however, live in 3D: a full 6-DoF pose sequence with metric translation and rotation. Such spatial trajectories are notoriously hard to obtain-- capturing them requires depth sensors, CAD models, multi-view rigs, or marker-based setups, and annotating them at scale is far more expensive than labeling 2D data.
In contrast, 2D observations and natural-language task descriptions are abundant and effortless to provide. This gap motivates the central question of this paper: \emph{can we generate dynamic 3D object motion trajectories from a single 2D image and an instruction alone?}

Existing methods fall into two broad families, and both leave key problems unresolved.
The first family \emph{learns dynamics directly from interaction data}: given observations of the scene, a feed-forward or diffusion model regresses the future object pose sequence~\citep{objectforesight,egoflow,egoscaler}. These methods inherit two limitations. \emph{(i) Data.} The egocentric manipulation corpora they rely on are either automatically pseudo-labeled at scale---trading label quality for quantity---or manually curated but small; moreover, their semantic annotations are coarse (e.g., verb--noun tags), which is insufficient for learning fine-grained, language-conditioned object motion. \emph{(ii) Privileged inputs.}
State-of-the-art predictors such as ObjectForesight~\citep{objectforesight} require multi-frame visual context together with the object's CAD mesh, while others additionally rely on depth input or the object's initial 6-DoF pose~\citep{egoscaler,egoflow}. 
Such requirements are rarely satisfied in the wild and severely restrict practical deployment. The second family \emph{outsources imagination to video generation}: RIGVid~\citep{rigvid} and its successors~\citep{gen2act,novaflow} prompt a video generator to synthesize a complete future video, then run a cascade of off-the-shelf perception modules, such as segmentation, point tracking, depth estimation, pose fitting, to extract a trajectory from the generated pixels. While this route does leverage the rich motion prior of large video models, 
accessing that prior through full video generation and pixel-space extraction incurs substantial inference cost.

We address these three problems head-on. \emph{First}, to remedy the shortage of high-quality supervision, we curate the \DATASET{} dataset, an object-centric egocentric manipulation corpus of \textbf{5,038 human-inspected 6-DoF object trajectories}. Instead of scaling up noisy automatic labels, every trajectory is manually inspected and paired with a fine-grained language instruction that describes the specific manipulation, providing the precise language-to-motion grounding that existing corpora lack. \emph{Second}, to eliminate the need for privileged inputs, we deliberately compress the interface to its minimum:
\METHOD{} takes \textbf{a single RGB frame and a task instruction}, no need for any other condition input, which makes it applicable in exactly the settings where prior methods break down.
\emph{Third}, instead of generating a full video with an external model and parsing its pixels, we generate trajectories \textbf{implicitly from the internal features of a locally hosted video diffusion model}. 
The key insight is that a clip-specific, decodable object-motion representation emerges in the intermediate features of an image-to-video diffusion model before the video is fully denoised or decoded. DreamTraj reads this latent motion signal and converts it into a relative 6-DoF trajectory, with metric scale restored using a depth estimate from the input frame.

Concretely, \METHOD{} feeds the input frame and instruction into a frozen image-to-video diffusion backbone and lets it denoise toward an imagined future. At an early denoising step we extract two complementary signals from the backbone: query--key attention maps, which implicitly track object points across the imagined frames, and pooled intermediate features, which carry scene-level geometry. A lightweight flow-matching readout head, conditioned on these signals, then decodes the future object trajectory as a sequence of relative 9-D pose tokens (3-D translation plus 6-D rotation), and a monocular depth estimate of the single input frame anchors the trajectory to metric scale. The backbone stays entirely frozen---dynamics knowledge is borrowed from internet-scale video pretraining rather than re-learned from our comparatively small interaction data---and because the readout happens at an early denoising step, no video is ever fully synthesized or decoded, which is what makes the trajectory readout several times faster than generating and parsing a full video.

Our contributions can be summarized as follows:
\begin{itemize}
\item \textbf{A object-centric trajectory dataset.}
We introduce the \DATASET{} dataset, \textbf{5,038 human-inspected} egocentric
6-DoF object trajectories paired with fine-grained language instructions,
providing high-quality supervision for language-conditioned object motion
generation.

\item \textbf{An implicit trajectory-generation paradigm.}
We propose the first approach that directly decodes object 6-DoF
trajectories from the internal representations of a frozen video diffusion
model. This formulation requires only a single RGB image and a task
instruction, without privileged geometric inputs, while avoiding explicit video generation and subsequent
pixel-space trajectory extraction.

\item \textbf{Accurate and efficient trajectory prediction.}
Extensive experiments show that \METHOD{} achieves state-of-the-art
translation and rotation prediction while being substantially more efficient
than generate-then-extract pipelines.
\end{itemize}

\section{Related Work}
\label{sec:related}

\subsection{Egocentric Manipulation Datasets}
Existing corpora trade annotation quality against accessibility. Large-scale
collections such as Ego4D~\citep{ego4d} and EPIC-KITCHENS~\citep{epickitchens}
offer thousands of hours of video but stop at verb--noun tags, with no
continuous object pose; precisely annotated
datasets~\citep{hoi4d,egoexo4d,hot3d,arctic} require RGB-D sensors, Aria
glasses, or lab rigs; and robot datasets~\citep{droid,oxe} log end-effector
actions, not object motion. Closest to us,
ObjectForesight~\citep{objectforesight} auto-mines millions of 6-DoF
trajectories from raw video, but inherits the mining pipeline's compounded
noise and only clip-level semantics. The \DATASET{} dataset targets this gap: its 5,038
trajectories are manually inspected and paired with fine-grained
instructions---the language-to-motion supervision the coarse and mined corpora
both lack.

\subsection{Object Trajectory Prediction}
Motion prediction as a manipulation interface almost always presumes
privileged input at inference time. Point-track and flow
methods~\citep{track2act,atm,generalflow,im2flow2act} need goal images, query
points, or RGB-D, and their 2D tracks still have to be lifted to 3D; 6-DoF
forecasters need video context, an object mesh, scene geometry, or an initial
pose~\citep{objectforesight,egoflow,egoscaler}, and SP-VTP~\citep{spvtp}
forecasts the end-effector from spatial prompts rather than language.
\METHOD{} commits to the minimal interface---one RGB frame and
a task instruction, with depth and the object's initial 3D position
\emph{estimated from the frame itself}. Only RIGVid~\citep{rigvid} shares
this interface, at the cost of full video synthesis and an external
perception cascade (Sec.~\ref{sec:related:prior}).

\subsection{Video Generation Models as Motion Priors}
\label{sec:related:prior}
Generate-then-extract pipelines tap video priors by synthesizing the future
and parsing its pixels: via inverse dynamics or dense
flow~\citep{unipi,avdc}, video-conditioned policies~\citep{dreamitate,gen2act},
or a generator--filter--pose-tracker cascade~\citep{rigvid,novaflow,genvid2robot}.
All of them denoise and decode a full video before any motion
is recovered. Yet the knowledge
these pipelines re-extract from pixels is already explicit \emph{inside}
diffusion models: image diffusion features carry semantic
correspondence~\citep{dift,sddino}, and video-diffusion attention supports
zero-shot point tracking~\citep{difftrack,ditracker,pointprompting,track4gen}
---but only on observed or fully generated videos. \METHOD{} reads
that plan out of the latent instead, without rendering a frame.

\section{Dataset Construction}
\label{sec:data}

Given the small capacity of the Reader and the frozen backbone, supervision \emph{quality} becomes the decisive factor, overshadowing sheer quantity. To this end, we curate the \textbf{\DATASET{}} dataset through a three-phase pipeline: selection, object-centric re-annotation, and generative augmentation. This process compresses 7,246 raw clips into 2,975 high-quality real trajectories, and after augmentation yields \textbf{5,038 final instances}.

\subsection{Selection and Annotation}
\label{sec:data:sources}%
\paragraph{Selection.}
We draw on six egocentric hand--object corpora---HOI4D~\citep{hoi4d},
TACO~\citep{taco}, HOT3D~\citep{hot3d}, H2O~\citep{h2o},
OakInk2~\citep{oakink2}, and FPHA~\citep{fpha}---each chosen for one reason: it
supplies frame-wise 6-DoF poses for a rigid manipulated object, the single
label our task cannot recover from RGB. Clips are unified into a camera-frame
representation: an object pose sequence $\{T_t\}$ relative to frame$_0$,
intrinsics $K$, and the mesh.

From the \textbf{7,246} pooled clips we keep \textbf{2,975}, admitting one only
if its interaction is \emph{meaningful} (a purposeful manipulation, not idle or
near-static motion a mean prior would already predict), \emph{describable} by a
single fine-grained instruction, and \emph{clean and complete} (object
groundable in the first frame, pose track gap-free over the horizon). Clips
that are otherwise good but bundle several actions, or contain a partial one,
are temporally cropped rather than discarded, so every retained clip holds
exactly one complete action.

\paragraph{Annotation.}
\label{sec:data:annot}%
The retained clips still carry only coarse action categories (``pick up'') that
name neither \emph{which} object moves nor \emph{how}. We re-annotate every
retained clip
from scratch: an AWQ-quantized \textbf{Qwen3.5-27B}~\citep{qwenvl} drafts an
object-centric label---manipulated object, verb, and manner of motion---from
sampled frames, and a human then corrects it against the video and fixes the
action's crop boundaries.
\subsection{Generative Augmentation}
\label{sec:data:twins}
The Reader runs on features of \emph{generated} videos, but trustworthy poses
exist only for real ones (Sec.~\ref{sec:method:twostage}), so we bridge the gap
by passing each retained clip's first frame and instruction to Wan2.2~\citep{wan};
fresh seeds give several plausible futures, which the pipeline below labels and
a human screens against the recovered track, yielding
\textbf{2,063} generated trajectories and bringing the \DATASET{} dataset to
\textbf{5,038}.
Since each inherits its source's object and instruction, augmentation
\emph{deepens} coverage rather than widening it: the corpus keeps the same
\textbf{14} object categories and \textbf{11} verbs (\textbf{92} object--verb
pairs) but packs more trajectories into each. Both halves are human-screened, but generated
poses come from our pipeline rather than a capture rig, so we keep them
separate and use the generated half only as the feature-alignment domain of
Sec.~\ref{sec:method:twostage}.

\subsection{Trajectory Extraction Pipeline}
\label{sec:data:pipeline}
\begin{figure}[t]
\centering
\includegraphics[width=\linewidth]{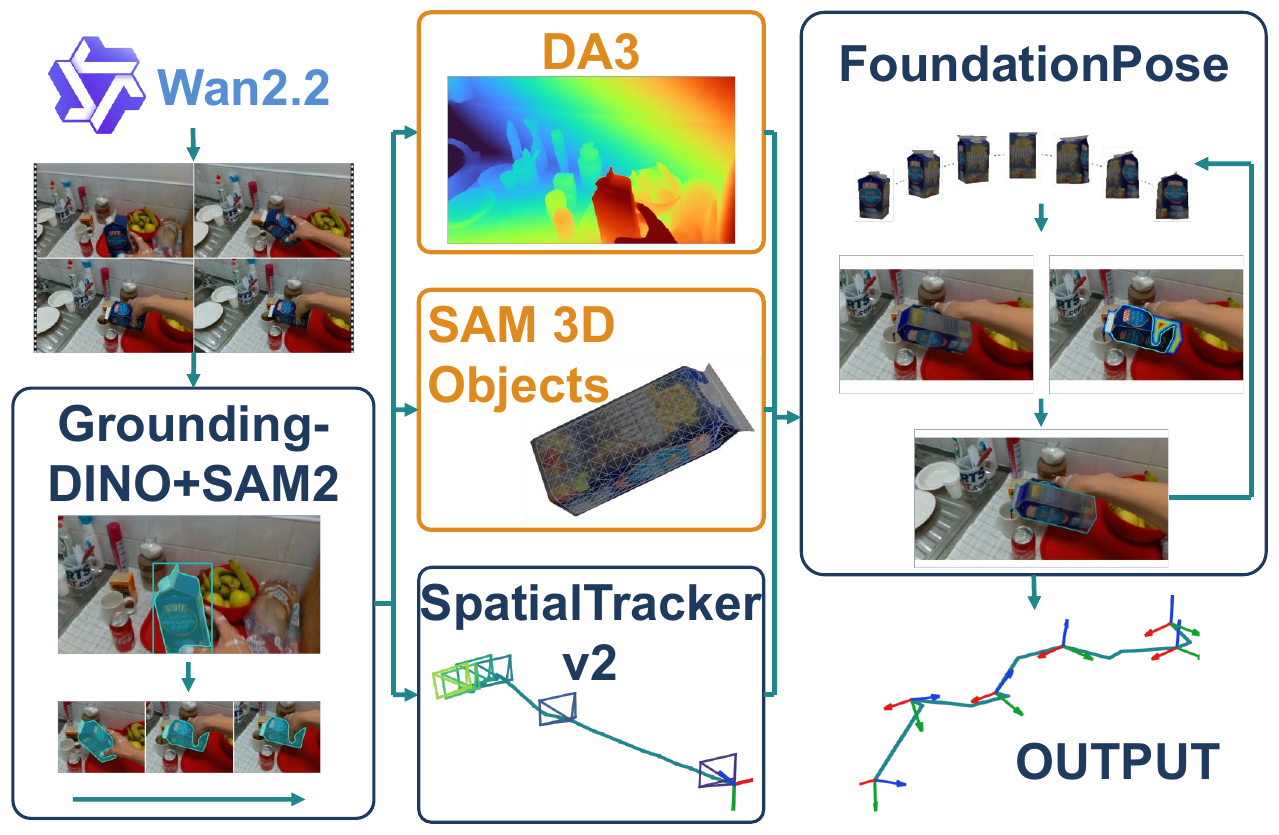}
\caption{\textbf{Trajectory extraction pipeline} used to label generated videos
(Sec.~\ref{sec:data:pipeline}). Amber marks the two blocks that depart from
RigVid and ObjectForesight: DA3 metric depth and a SAM 3D Objects mesh.}
\label{fig:pipeline}
\end{figure}

To label a generated video, we recover a metric 6-DoF trajectory from RGB alone
(Fig.~\ref{fig:pipeline}) following RigVid~\citep{rigvid} and
ObjectForesight~\citep{objectforesight}: we ground the object in the first frame
(GroundingDINO~\citep{groundingdino}~+~SAM2~\citep{sam2}) and track its per-frame
pose by render-and-compare with FoundationPose~\citep{foundationpose}, given the
object mesh, metric depth, and camera geometry from
SpatialTracker~v2~\citep{spatialtrackerv2}. Two changes are necessary for
generated content: metric depth from DA3~\citep{da3} rather than a
relative-depth predictor, removing the per-clip scale ambiguity that would
corrupt every translation label; and the object mesh reconstructed by SAM 3D
Objects~\citep{sam3dobjects} rather than a CAD model, which for much of our data
is unavailable or too imprecise to rely on.

\section{Method}
\label{sec:method}

\begin{figure*}[t]
\centering
\includegraphics[width=\linewidth]{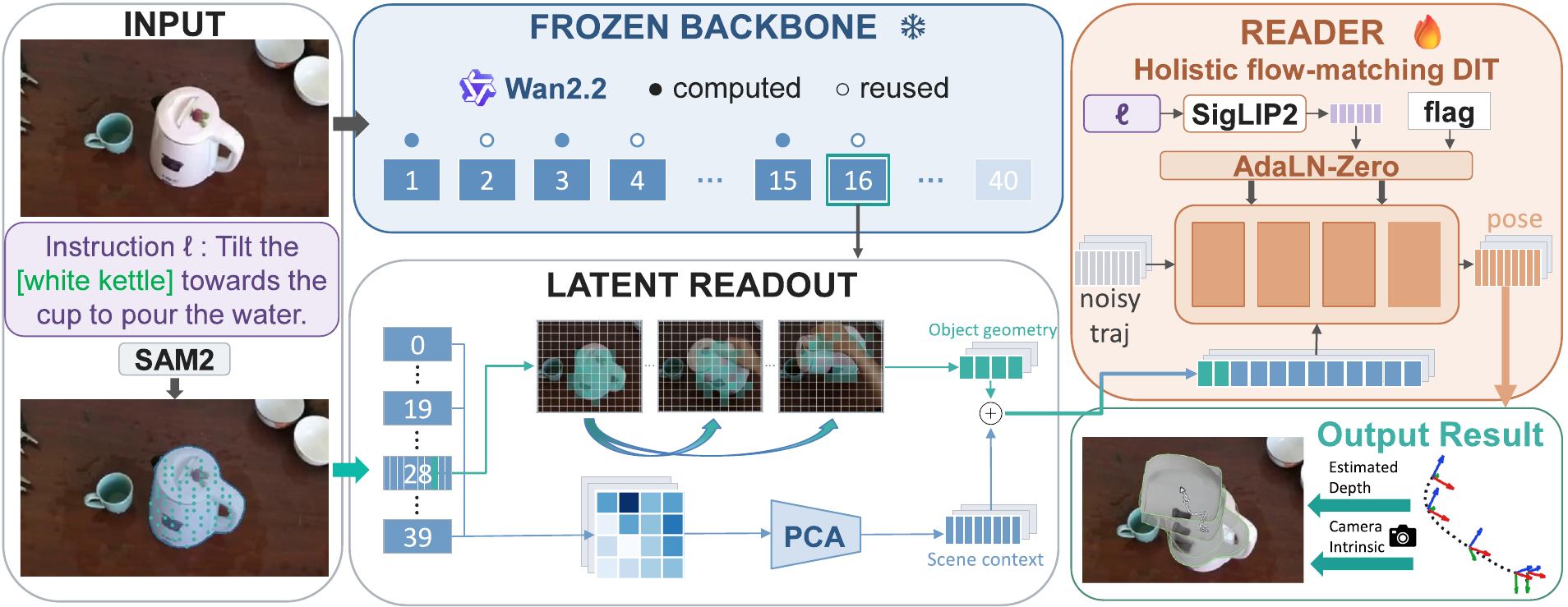}
\caption{\textbf{Overview of \METHOD{}.} One RGB frame and one instruction go
in. The object is grounded once in that frame (GroundingDINO~+~SAM2) to give
query points; a \emph{frozen} Wan2.2-I2V-A14B is stopped at step $16$ of $40$,
and from that latent we read q$\cdot$k attention tracks (block $28$, head $32$)
and anchor-pooled hidden states (blocks $0/19/39$) as a $13\times132$ sequence.
A $7.1$M-parameter flow-matching trainable Reader, turns it
into $13$ relative 9-D pose tokens, which, with a depth anchor $z_0$ and the intrinsics $K$ lift to metric 6-DoF.}
\label{fig:overview}
\end{figure*}

\METHOD{} takes a single RGB frame $I_0$ and a language instruction $\ell$ and
returns the manipulated object's 6-DoF trajectory over the next $49$ frames
($\sim$2\,s), in the camera frame of $I_0$ (Fig.~\ref{fig:overview}). The design
splits into a \emph{frozen} backbone that supplies dynamics and a small
\emph{trained} Reader that decodes them, and unfolds in three stages: the backbone
imagines a plausible future, forming a conditional motion hypothesis, and we read two motion signals from its latent, q$\cdot$k attention tracks and pooled hidden states
(Sec.~\ref{sec:method:backbone}); the Reader denoises these into relative pose
tokens (Sec.~\ref{sec:method:reader}); and it is trained across the real and
generated feature domains it meets at deployment (Sec.~\ref{sec:method:twostage}).
A single depth estimate lifts the scale-free output to metric units.

We represent the trajectory as frame-$0$-anchored relative poses.
For latent frame $t$ the Reader emits a $9$-D token
\begin{equation}
\label{eq:token}
  \mathbf{y}_t \;=\;
  \big[\,\underbrace{\delta u_t,\ \delta v_t}_{\text{bearing}},\;
        \underbrace{s_t}_{\log\text{-depth}},\;
        \underbrace{\mathbf{r}_t \in \mathbb{R}^6}_{\text{rotation}}\,\big],
\end{equation}
where $(\delta u_t, \delta v_t)$ is the object's displacement from its frame-$0$
bearing in normalised image coordinates, $s_t = \log(z_t/z_0)$ its log-depth
ratio, and $\mathbf{r}_t$ the continuous 6-D representation~\citep{rot6d} of the
relative rotation $R_0^{\!\top}R_t$. Every channel is a difference or ratio, so the parameterisation is
scale-free: the Reader is never asked for absolute scale; metric units are
restored only at the output (Sec.~\ref{sec:method:reader}) from one monocular
depth estimate.

\subsection{Reading Motion from a Frozen Video Prior}
\label{sec:method:backbone}
\label{sec:method:readout}
We use Wan2.2~\citep{wan}, an image-to-video diffusion transformer, kept entirely
frozen. Conditioned on $I_0$ and the instruction $\ell$, it denoises toward an
imagined future, and we read its internal features at a single early step---step
$16$ of the $40$-step schedule ($t\!\approx\!882$, in Wan's low-noise expert),
where the read-out motion signal saturates (Sec.~\ref{sec:exp:readout}). To read
faster we apply stride-2 flow caching---recomputing the guidance-combined flow
every other step and reusing it in between---which roughly halves the backbone
forwards (Sec.~\ref{sec:exp:eff}).

These features carry two complementary signals. For object motion we treat
query--key attention as an implicit point tracker~\citep{difftrack,ditracker,dift}:
we ground the object once in $I_0$ (GroundingDINO~\citep{groundingdino} and
SAM2~\citep{sam2}) to get query patches, then soft-match each against every patch
of every frame by head-averaged cosine similarity, always against frame $0$ rather
than chaining, which trades tolerance of appearance change for freedom from drift.
This yields a soft track of the object through the imagined future, from which we
read three per-frame quantities (Eq.~\ref{eq:token}): the \emph{centroid} (a 2-D
bearing), the \emph{spread} $\sigma_f$ (a relative-depth cue, since apparent size
scales as $1/z$, so $\log(\sigma_0/\sigma_f)$ tracks $s_f$), and the
\emph{in-plane rotation} (closed-form from the $2\times2$ cross-covariance of
frames $0$ and $f$, whose sign is stable under a polar decomposition)---four
scalars in all, read from one (block, head) cell fixed on training folds
(block $28$, head $32$; Sec.~\ref{sec:exp:readout}).

The second signal is scene context: we mean-pool hidden states from blocks $0$,
$19$, $39$ over a $4\times4$ anchor grid and project them, via an in-fold PCA, to
$128$ dimensions. The Reader thus receives a sequence of $13$ vectors of
$128{+}4=132$ dimensions---$128$ of scene context beside $4$ of object geometry.

\subsection{Flow-Matching Trajectory Reader}
\label{sec:method:reader}

The readout of Sec.~\ref{sec:method:backbone} is a coarse motion signal---a 2-D
object track with per-frame depth and rotation cues, beside scene context. A small
\emph{holistic} flow-matching DiT~\citep{dit,flowmatching} decodes it into the
final trajectory, denoising all $13$ pose tokens at once so that global properties
such as total displacement are decided jointly rather than accumulated. It has
$7.1$M trainable parameters, three orders of magnitude below the backbone---it
only has to decode the motion the backbone already committed to.

The Reader takes three inputs. The $132$-D readout (Sec.~\ref{sec:method:backbone})
is projected to $13$ context tokens (width $256$) and supplied as cross-attention
\emph{memory}, so the trajectory can attend to the read-out motion frame by frame;
the noisy trajectory $\mathbf{x}_\tau\in\mathbb{R}^{13\times9}$ is the token stream
being denoised; and the clip-level conditioning---denoising time $\tau$, a $768$-D
SigLIP2 embedding of the instruction $\ell$, and a two-way domain flag $d$
(Sec.~\ref{sec:method:twostage})---is summed, each through its own encoder, into an
AdaLN-Zero modulation vector
\begin{equation}
\label{eq:cond}
  \mathbf{c} \;=\; \bar{\mathbf{z}}_{\text{ctx}}
   \;+\; \mathrm{emb}(\tau)
   \;+\; E_{\text{txt}}(\mathbf{e}_\ell)
   \;+\; E_{\text{dom}}(d).
\end{equation}
The body stacks four DiT blocks, each applying self-attention over the pose tokens,
cross-attention into the context tokens, and an MLP, all gated by $\mathbf{c}$; a
zero-initialised head then emits the $9$-D flow-matching velocity per token---the
$13\times9$ output. We train with conditional flow matching---pinning frame $0$ to
the anchor and adding light endpoint, SO(3)-geodesic, and smoothness terms---and at
inference integrate $20$ Euler steps to produce the trajectory.

Finally we map the $13$ scale-free tokens to a metric trajectory: we resample them
to all $49$ frames and fix absolute scale from the object's depth in the input
frame---a sensor reading when one is available, otherwise a monocular
estimate~\citep{depthanything}---after which the camera intrinsics recover metric
position. Depth enters only here, at the output.

\subsection{Training across Two Feature Domains}
\label{sec:method:twostage}

The Reader's features come, at deployment, from a \emph{generated} video, yet
trustworthy 6-DoF supervision exists only for \emph{recorded} video: training on
recorded features alone leaves a feature-distribution gap at deployment, while
training on generated features alone inherits pipeline-label noise. We therefore
train on both with one shared weight set and let the domain flag $d$
(Eq.~\ref{eq:cond}) absorb the difference. In \emph{domain~A} we noise a recorded
video to the read step, run one backbone forward, and extract the features of
Sec.~\ref{sec:method:readout}, supervised by the corpus's true camera-frame poses.
In \emph{domain~B} the backbone instead imagines a future from the same first
frame and instruction, and we supervise against \emph{that generated video's own
motion} (Sec.~\ref{sec:data:pipeline})---never the recorded trajectory, since the
backbone commits to \emph{a} plausible motion, not \emph{the} one that happened.
This keeps the generated domain honest and aligns training with deployment. To make the mixture work we warm up on domain~B
before mixing in the real labels, and additionally down-weight B's noisier
channels.

\section{Experiments}
\label{sec:exp}

\subsection{Experimental Setup}
\label{sec:exp:setup}

\paragraph{Data.}
All experiments use the \DATASET{} dataset, whose sources, admission criteria,
filters, and verification protocol are detailed in Sec.~\ref{sec:data}. We
evaluate with 5-fold cross-validation over the $5{,}038$ samples, with folds
assigned \emph{by source clip}: a recorded clip and every generated clip derived
from it share a fold, so no generated twin of a held-out clip is ever seen in
training.

\paragraph{Metrics.}
\providecommand{\transunit}{metres}
Translation is scored by displacement error in \transunit, averaged over the $13$
pose tokens of a clip (ADE) and at the last token (FDE); displacements are
taken relative to the anchor frame, so the comparison is independent of the
reference frame. Rotation is scored by the SO(3) geodesic angle
$\theta = \arccos\!\big((\mathrm{tr}(R_\text{pred}^{\!\top} R_\text{gt}) - 1)/2\big)$,
in degrees, again averaged over the clip (Rot) and at the last token
(Rot-final). Errors are per-clip medians throughout. Over all $5{,}038$
out-of-fold samples the deployed Reader attains $6.4$\,cm ADE, $9.5$\,cm FDE
and $24.6^\circ$ rotation error; the comparisons below are each restricted to
the protocol they name.

\paragraph{Baselines.}
We compare \emph{accuracy} against two prior 6-DoF forecasters,
ObjectForesight~\citep{objectforesight} and EgoScaler~\citep{egoscaler}
(Table~\ref{tab:baselines}), and \emph{inference cost} against a RIGVid-style
generate-then-extract cascade~\citep{rigvid,dream2flow,novaflow}
(Table~\ref{tab:efficiency}), which we instantiate ourselves on the same
backbone and run on the same generated videos as \METHOD{}.
Both baselines are retrained on the \DATASET{} dataset, EgoScaler starting from
its released 7B checkpoint. Both consume inputs \METHOD{} does not, and we mark them
$\dagger$: ObjectForesight~\citep{objectforesight} takes video context, the
object's mesh, and three ground-truth context poses;
EgoScaler~\citep{egoscaler} takes depth and the object's initial pose.
\METHOD{} sees one RGB frame and the instruction, estimating depth and the
object's position from that frame itself.

\paragraph{Implementation.}
The Reader is the only trained component; the backbone is frozen throughout.
All five cross-validation folds fit in $16$ minutes on one RTX PRO 6000.
Training features are extracted under the same stride-2 caching used at
inference, so the Reader sees the same computation in both.
\subsection{Comparison with Prior Forecasters}
\label{sec:exp:baselines}
\METHOD{} predicts over a window longer than either baseline covers, so in each
block we adapt \METHOD{}'s inference window to that baseline---resampling onto
its timestamps and horizon---rather than the reverse.

\METHOD{} leads EgoScaler on every metric, despite that baseline receiving
depth and the object's initial pose. The lead survives both
stress tests we ran on it:
sweeping five input conventions for the baseline and taking its
\emph{best} value per metric, and scoring on our native token timestamps
instead of its own grid.

ObjectForesight is scored in its own block, its horizon fixed by its context
requirement. Retrained on the \DATASET{} dataset it trails \METHOD{} on translation and
rotation alike.

\begin{table}[!t]
\centering
\small
\setlength{\tabcolsep}{1.1mm}
\begin{tabular}{@{}lcccc@{}}
\toprule
Method & ADE$\downarrow$ & FDE$\downarrow$ & Rot$\downarrow$ & Rot-final$\downarrow$ \\
\midrule
\multicolumn{5}{@{}l}{\emph{EgoScaler protocol} --- its native $2.0$\,s window, 20 timestamps} \\
\rowcolor[gray]{0.92} \multicolumn{5}{@{}l}{\;\;(a) real capture, real GT ($n{=}537$)} \\
EgoScaler$^\dagger$ & $7.52$ & $10.71$ & $28.6$ & $45.6$ \\
\METHOD{} & $\mathbf{3.04}$ & $\mathbf{4.71}$ & $\mathbf{7.9}$ & $\mathbf{12.9}$ \\
\rowcolor[gray]{0.92} \multicolumn{5}{@{}l}{\;\;(b) generated video, pipeline GT ($n{=}414$)} \\
EgoScaler$^\dagger$ & $6.55$ & $9.52$ & $31.7$ & $50.0$ \\
\METHOD{}                            & $\mathbf{3.01}$ & $\mathbf{4.72}$ & $\mathbf{11.5}$ & $\mathbf{18.3}$ \\
\midrule
\multicolumn{5}{@{}l}{\emph{ObjectForesight} --- its native $1.17$\,s horizon} \\
\rowcolor[gray]{0.92} \multicolumn{5}{@{}l}{\;\;(a) real capture, real GT ($n{=}593$)} \\
ObjectForesight$^\dagger$ & $2.71$ & $4.58$ & $7.8$ & $11.1$ \\
\METHOD{} & $\mathbf{1.97}$ & $\mathbf{3.32}$ & $\mathbf{6.5}$ & $\mathbf{10.5}$ \\
\rowcolor[gray]{0.92} \multicolumn{5}{@{}l}{\;\;(b) generated video, pipeline GT ($n{=}414$)} \\
ObjectForesight$^\dagger$ & $2.80$ & $4.42$ & $8.6$ & $14.7$ \\
\METHOD{}                                        & $\mathbf{1.97}$ & $\mathbf{3.20}$ & $\mathbf{7.1}$ & $\mathbf{11.6}$ \\
\midrule
\multicolumn{5}{@{}l}{\emph{ObjectForesight} --- extended to the whole action} \\
\rowcolor[gray]{0.92} \multicolumn{5}{@{}l}{\;\;(a) real capture, real GT ($n{=}594$)} \\
ObjectForesight$^\dagger$ & $10.29$ & $12.74$ & $29.8$ & $30.2$ \\
\METHOD{} & $\mathbf{6.76}$ & $\mathbf{8.39}$ & $\mathbf{18.2}$ & $\mathbf{20.4}$ \\
\rowcolor[gray]{0.92} \multicolumn{5}{@{}l}{\;\;(b) generated video, pipeline GT ($n{=}366$)} \\
ObjectForesight$^\dagger$ & $10.26$ & $13.51$ & $37.3$ & $51.0$ \\
\METHOD{}                                        & $\mathbf{7.26}$ & $\mathbf{9.34}$ & $\mathbf{33.0}$ & $\mathbf{41.4}$ \\
\bottomrule
\end{tabular}
\caption{\textbf{Main comparison} on held-out folds. Block (a) restricts to
clips captured by a real sensor, so the reference trajectory is the dataset's
own 6-DoF annotation; block (b) is the full evaluation set, whose labels come
from the extraction pipeline of Sec.~\ref{sec:data:pipeline}.
ADE/FDE in cm, rotation in degrees. Baselines marked $\dagger$
consume inputs \METHOD{} does not (Sec.~\ref{sec:exp:setup}). Each baseline predicts on
its own timestamps---EgoScaler a fixed 20-step trajectory, ObjectForesight a
horizon fixed by its context requirement---so each is reported against
\METHOD{} evaluated at \emph{that} baseline's timestamps, and blocks are not
comparable to one another.}
\label{tab:baselines}
\end{table}

\begin{figure*}[!t]
\centering
\includegraphics[width=\linewidth]{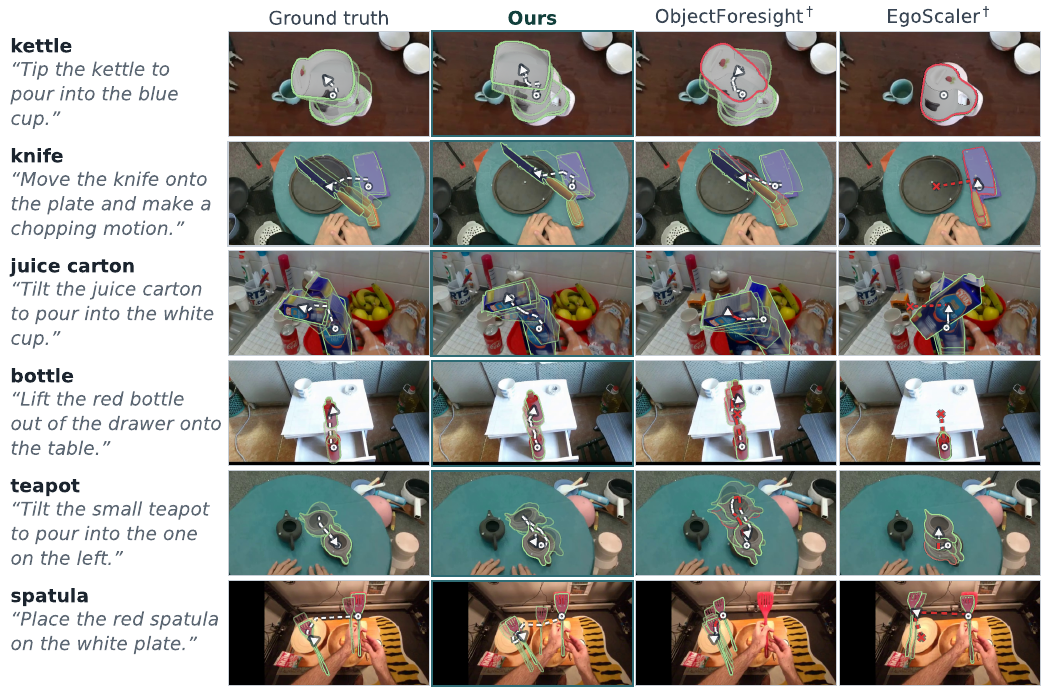}
\caption{\textbf{Qualitative comparison.} Each row is one clip, labelled with
the instruction the model was given (abridged for space). Panels overlay $7$ of
the $13$ predicted poses on the anchor frame, each outlined in green, with the
dashed line tracing the object's centre.
Red marks error: a centroid off by more
than the object's own width, or an orientation off by more than $45^\circ$.
Methods marked
$\dagger$ receive privileged input: ObjectForesight is
given the first three poses as ground-truth context, so its overlay starts from
the correct pose by construction. EgoScaler predicts over its native $2$\,s
horizon, the other columns over the whole action.}
\label{fig:compare}
\end{figure*}

\begin{figure*}[!t]
\centering
\includegraphics[width=\linewidth]{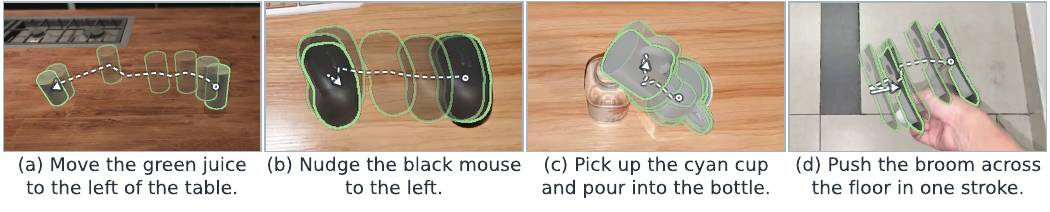}
\caption{\textbf{Generalization beyond the training distribution.} Scenes,
objects and actions absent from every training corpus, rendered as in
Fig.~\ref{fig:compare}. Panel~(a) is a video-game frame; (b--d) are phone
photographs. Instructions abridged.}
\label{fig:wild}
\end{figure*}

\paragraph{Qualitative comparison.} Fig.~\ref{fig:compare} overlays predicted
poses of a clip onto its anchor frame, rendering the object's own reconstructed
mesh at each pose so the trajectory is read as motion of the object rather than
as an abstract curve. %
Each pose is outlined so it separates from the scene, and the object's centre is
traced, making the final placement and the route taken directly comparable
across methods.

\subsection{Efficiency}
\label{sec:exp:eff}
Table~\ref{tab:efficiency} compares end-to-end inference cost against the
generate-then-extract route~\citep{rigvid,dream2flow,novaflow} on identical
hardware (one RTX PRO 6000, exclusive). That route must run the full denoising
schedule, decode the video, and execute a four-model perception stack
(segmentation, point tracking, monocular depth, pose fitting). We measure its
generation stage at $280$\,s per clip for $80$ guided forwards plus VAE
decoding, and its perception stack at $68$\,s per clip, both averaged over
three clips on one GPU. \METHOD{} stops at denoising step $k{=}16$ of $40$ and
runs no external perception: $76$\,s per 49-frame trajectory, a $4.6\times$
end-to-end speedup.

\begin{table}[!t]
\centering
\small
\setlength{\tabcolsep}{0.55mm}   %
\begin{tabular}{@{}lccccc@{}}
\toprule
Method & Fwd. & Decode & Percep. & Generate & Total$\downarrow$ \\
\midrule
Generate-then-extract & $80/80$ & yes & 4 models & $280$\,s & $348$\,s \\
\METHOD{}             & $\mathbf{18/80}$ & no & none & $\mathbf{76}$\,\textbf{s} & $\mathbf{76}$\,\textbf{s} \\
\bottomrule
\end{tabular}
\caption{\textbf{Inference cost} vs.\ the generate-then-extract route,
identical hardware. Guided forwards counts both classifier-free-guidance
branches per denoising step. All times are measured on one GPU, averaged over three clips.}
\label{tab:efficiency}
\end{table}

\begin{table}[!t]
\centering
\small
\begin{tabular}{@{}lcccccc@{}}
\toprule
Read step $k$ & $12$ & $14$ & & $16$ & $18$ & $20$ \\
Expert        & \multicolumn{2}{c}{high-noise} & & \multicolumn{3}{c}{low-noise} \\
\midrule
Track margin$\uparrow$   & $0.123$ & $0.129$ & & $\mathbf{0.191}$ & $0.196$ & $0.197$ \\
ADE (cm)$\downarrow$     & $8.05$  & $8.52$  & & $\mathbf{7.89}$  & $8.20$  & $8.29$ \\
\bottomrule
\end{tabular}
\caption{\textbf{Readout step.} \emph{Track margin} is the permutation-controlled score of
Eq.~\ref{eq:margin} between the attention centroid and the recorded 2D object
track ($955$ clips). \emph{ADE} comes from retraining the Reader at each step on
the generated-video path ($321$ clips); its scale is not comparable to
Table~\ref{tab:baselines}. Bold marks the adopted step, the earliest at which
the margin saturates.}
\label{tab:readstep}
\end{table}

\begin{table}[!t]
\centering
\small
\setlength{\tabcolsep}{0.9mm}   %
\begin{tabular}{@{}lcccc@{}}
\toprule
Variant & ADE$\downarrow$ & FDE$\downarrow$ & Rot$\downarrow$ & Depth corr.$\uparrow$ \\
\midrule
Control (deployed)           & $\mathbf{6.42}$ & $\mathbf{9.54}$ & $\mathbf{24.63}$ & $\mathbf{0.640}$ \\
w/o pooled hidden states     & $6.91$ & $10.47$ & $26.00$ & $0.565$ \\
w/o q$\cdot$k track          & $7.13$ & $9.90$ & $26.15$ & $0.637$ \\
w/o instruction conditioning & $6.45$ & $9.70$ & $26.20$ & $0.639$ \\
\bottomrule
\end{tabular}
\caption{\textbf{Ablations.} ADE/FDE in cm, rotation in degrees, over all
$5{,}038$ out-of-fold samples. All four arms are the deployed configuration
retrained from the same features, differing only in the conditioning removed.
\emph{Depth corr.} is the median over clips of $|r|$ between the predicted and
ground-truth log-depth ratio $s_t=\log(z_t/z_0)$, each linearly detrended
against frame index.}
\label{tab:ablations}
\end{table}

\subsection{Where to Read: Head and Step}
\label{sec:exp:readout}
Two choices decide what the Reader sees: \emph{which} attention head supplies
the object track, and at \emph{which} denoising step it is read. We settle both
with a permutation-controlled \emph{margin} between the attention centroid and
the recorded 2D track,
\begin{equation}
\label{eq:margin}
  m \;=\; \big|\rho(a_i, g_i)\big| \;-\;
          \mathbb{E}_{j \neq i}\,\big|\rho(a_i, g_j)\big| ,
\end{equation}
where $a_i$ is the readout on clip $i$ and $g_i$ its ground truth, both linearly
detrended against frame index: correlation against the clip's own motion, minus
the score the same track achieves against other clips'---any shared population
prior cancels, so $m\!>\!0$ certifies per-sample signal. The control is not a
formality: these trajectories share a strong temporal profile, and a readout
locked onto that profile alone still correlates $0.43$ with an \emph{unrelated}
clip's ground truth.
\vspace{-1em}

\paragraph{Head Selection.}
A pass over all $40\times40$ (block, head) pairs puts block $28$, head $32$ on
top (margin $0.191$ over $955$ clips), and we read from that cell.

\paragraph{Step Selection.}
Sweeping $k$ for that cell gives a discontinuous answer
(Table~\ref{tab:readstep}): the margin gains $48\%$ from $k{=}14$ to $k{=}16$
(paired $t{=}12.6$ over $955$ clips), then stops moving, every later step adding
under $3\%$. The jump falls exactly on Wan's high- to low-noise expert switch
($t\!=\!900$, between steps $14$ and $16$), and $k{=}16$ is the earliest step
past it.

\subsection{Generalization Experiments}
\label{sec:exp:gen}
A frozen internet-scale prior should carry outside the data the Reader was
trained on. We collect $50$ scenes absent from every training corpus---$40$
real-captured and $10$ from a commercial video game---each paired with one
instruction, and run them in the deployed condition: one image in, one metric
6-DoF trajectory out. No trustworthy ground truth exists here, so
we score the split as a user would: ten annotators view the predicted trajectory
rendered as the object's own mesh swept along the predicted poses
(Fig.~\ref{fig:wild}) and judge whether the object's motion carries out the
instruction. \METHOD{} succeeds on $80\%$ of them.

\subsection{Ablation Studies}
\label{sec:exp:ablations}
Table~\ref{tab:ablations} removes one input at a time from the deployed
configuration and retrains. Every removal costs accuracy, and each leaves a
different signature. The q$\cdot$k track carries the trajectory's position:
without it ADE degrades the most of any arm ($+0.71$\,cm), as it is the only
input that follows the object itself rather than the scene around it. The
anchor-pooled hidden states carry the geometry that turns a track into a metric
displacement, so dropping them costs the endpoint and the depth correlation the
most. Dropping the instruction shifts the emphasis to rotation, consistent with
a given path admitting several rotations that only the language distinguishes;
that arm removes the instruction from \emph{the Reader alone}, since the
features it reads come from a denoising pass that was itself conditioned on
$\ell$.

\section{Conclusion}
\label{sec:conclusion}

In this paper, we presented \METHOD{}, a framework for predicting 6-DoF object trajectories from a single RGB frame and a language instruction by directly reading motion from the intermediate representations of a frozen image-to-video diffusion model, thereby bypassing explicit video synthesis.
To supply the supervision the task lacks, we built
the \DATASET{} dataset, $5{,}038$ object-centric egocentric trajectories paired with
fine-grained instructions. \METHOD{} beats prior forecasters on translation and
rotation alike with strictly less input, and runs $4.6\times$ faster than the
generate-then-extract pipeline whose labels it learns from. Driving a real
manipulator from the predicted trajectory is the natural next step.
\providecommand{\mainonly}[1]{#1}
\mainonly{These results show that video diffusion latents can serve as compact and directly decodable motion priors for embodied prediction.}

\bibliography{references}

\clearpage
\suppressfloats[t]
\appendix
\section*{Technical Appendix}
\noindent
This appendix reports the full experimental configuration
(Sec.~\ref{sec:app:config}), the cross-validation protocol
(Sec.~\ref{sec:app:folds}), baseline reproduction details
(Sec.~\ref{sec:app:baselines}), dataset statistics (Sec.~\ref{sec:app:data}),
supporting ablations (Sec.~\ref{sec:app:ablation}), and additional qualitative
results (Sec.~\ref{sec:app:qual}).

\section{Full Experimental Configuration}
\label{sec:app:config}

Sec.~5.1 of the main paper gives only the facts needed to read the tables.
This section reports every setting required to reproduce them.

\subsection{Backbone and Readout}

The backbone is Wan2.2-I2V-A14B, a dual-expert mixture-of-experts image-to-video
diffusion transformer. Table~\ref{tab:app:backbone} lists the generation and
readout settings. The
sampler, guidance scale and seed are held fixed across every experiment in the
paper, so any variation between runs comes from Reader training alone.

\begin{table}[t]
\centering
\small
\setlength{\tabcolsep}{1.2mm}
\begin{tabular}{@{}ll@{}}
\toprule
Setting & Value \\
\midrule
Backbone & Wan2.2-I2V-A14B (frozen) \\
Blocks / heads / width & $40$ / $40$ / $5120$ \\
Video resolution & $832 \times 480$ \\
Frames & $49$ @ $16$\,fps \\
VAE stride & $(4, 8, 8)$; patch $(1, 2, 2)$ \\
Latent grid & $13 \times 60 \times 104$ \\
Token grid & $13 \times 30 \times 52 = 20{,}280$ \\
\midrule
Sampler & FlowUniPCMultistep \\
Denoising steps & $40$ \\
Shift & $5.0$ \\
Guidance scale & $3.5$ \\
Seed & $0$ \\
\midrule
Readout step & $16$ of $40$ ($t \approx 882$) \\
Flow caching & stride-2 (compute one step, reuse one) \\
q$\cdot$k cell & block $28$, head $32$ \\
Pooled-hidden blocks & $0$, $19$, $39$ \\
Anchor grid & $4 \times 4$, anchor-mean pooled \\
PCA dimension & $128$ (over the concatenated $3 \times 5120$) \\
Reader input & $13$ tokens $\times$ ($128 + 4$) $= 132$\,D \\
\bottomrule
\end{tabular}
\caption{\textbf{Backbone and readout configuration.}}
\label{tab:app:backbone}
\end{table}

The four object-geometry channels read from the q$\cdot$k cell are the two
centroid coordinates (frame-$0$-relative, normalised by image resolution), the
log-spread ratio $\log(\sigma_0/\sigma_f)$, and the in-plane rotation angle
normalised by $90^\circ$.

\subsection{Reader Architecture}

The Reader is the only trained component. It is a holistic flow-matching DiT
that denoises all $13$ pose tokens jointly; Table~\ref{tab:app:reader} gives its
dimensions. Of the $7.08$M parameters, $6.88$M belong to the DiT body and
$0.20$M to the instruction encoder that maps the $768$-D SigLIP2 embedding into
the AdaLN-Zero modulation vector.

\begin{table}[t]
\centering
\small
\setlength{\tabcolsep}{1.2mm}
\begin{tabular}{@{}ll@{}}
\toprule
Setting & Value \\
\midrule
Width $d$ & $256$ \\
Blocks & $4$ \\
Attention heads & $8$ \\
MLP ratio & $4.0$ \\
Pose tokens & $13$ \\
Token dimension & $9$ ($2$ bearing $+$ $1$ log-depth $+$ $6$ rotation) \\
Conditioning & AdaLN-Zero, three gates per block \\
Context memory & $13 \times 256$, cross-attention \\
Extra conditions & instruction $768$\,D; domain flag $2$\,D \\
Positional encoding & fixed sinusoidal (pose and context) \\
Output head & zero-initialised linear \\
\midrule
Parameters & $7{,}078{,}165$ ($7.1$M) \\
Inference & $20$ explicit Euler steps \\
\bottomrule
\end{tabular}
\caption{\textbf{Reader architecture.}}
\label{tab:app:reader}
\end{table}

\subsection{Optimisation}

Table~\ref{tab:app:optim} lists the training schedule; the budget is fixed
rather than early-stopped, so no reported number depends on a validation-based
stopping rule. One epoch is $29$ optimiser steps at batch size
$128$; the deployed Reader is the snapshot at $5{,}799$ steps, i.e.\ the end of
epoch $200$. The ablation arms of Table~4 in the main paper use this same
schedule.

The training objective is conditional flow matching on the velocity field, with
three auxiliary terms: an endpoint term on the reconstructed final token
(weight $0.1$), an SO(3) chordal surrogate on the reconstructed rotations
(weight $0.1$), and a second-difference smoothness term on the translation
channels (weight $0.05$). Frame $0$ is pinned to the anchor pose in both the
noise and the target, and is excluded from the loss.

\begin{table}[t]
\centering
\small
\setlength{\tabcolsep}{1.2mm}
\begin{tabular}{@{}ll@{}}
\toprule
Setting & Value \\
\midrule
Optimiser & AdamW \\
Learning rate & $2 \times 10^{-4}$ (constant) \\
Betas & $(0.9, 0.95)$ \\
Weight decay & $0.01$ \\
Batch size & $128$ \\
Steps per epoch & $29$ \\
Deployed snapshot & $5{,}799$ steps (epoch $200$) \\
Stage-1 warm-up & $40$ epochs, domain B only \\
Stage-2 mixture & $60\%$ domain A per batch \\
Early stopping & none (fixed budget) \\
Precision & fp32 \\
\midrule
Flow-matching loss & velocity MSE \\
Endpoint term & $0.1$ \\
SO(3) chordal term & $0.1$ \\
Smoothness term & $0.05$ \\
\bottomrule
\end{tabular}
\caption{\textbf{Reader optimisation.} Identical across every arm of
Table~4 in the main paper; the arms differ only in which conditioning
signal is removed.}
\label{tab:app:optim}
\end{table}

\subsection{Metric Definitions}
\label{sec:app:metrics}

All errors are \emph{per-clip medians} over the evaluated set, not means, so a
handful of catastrophic clips cannot dominate a column. Translation is scored
after mapping the scale-free tokens to metric units.

\paragraph{ADE / FDE.}
Let $\hat{\mathbf{p}}_t, \mathbf{p}_t \in \mathbb{R}^3$ be the predicted and
ground-truth object positions at pose token $t$, both taken relative to the
anchor frame. Then
\begin{equation}
\mathrm{ADE} = \frac{1}{T}\sum_{t=1}^{T} \lVert \hat{\mathbf{p}}_t - \mathbf{p}_t \rVert_2,
\qquad
\mathrm{FDE} = \lVert \hat{\mathbf{p}}_T - \mathbf{p}_T \rVert_2 ,
\end{equation}
in centimetres. Because both sequences are anchor-relative, the comparison does
not depend on the choice of reference frame.

\paragraph{Rotation.}
Rotation error is the SO(3) geodesic angle in degrees,
\begin{equation}
\theta_t = \arccos\!\Big(\big(\mathrm{tr}(R^{\top}_{\text{pred},t} R_{\text{gt},t}) - 1\big)/2\Big),
\end{equation}
averaged over the clip (Rot) and taken at the last token (Rot-final).

\paragraph{Depth correlation.}
The depth channel is the log-depth ratio $s_t = \log(z_t / z_0)$, which is
scale-free by construction. A raw correlation between predicted and
ground-truth $s_t$ would be dominated by the shared monotone drift of an object
moving steadily toward or away from the camera, so we first remove that drift:
both the predicted and the ground-truth $s_t$ are \emph{linearly detrended
against frame index}, and the metric is the absolute Pearson correlation of the
two residuals, reported as the median over clips. It therefore measures whether
the predicted depth profile has the right \emph{shape}, independently of its
overall slope and offset.

\paragraph{Token count.}
Unless a table states otherwise, averages run over all $13$ pose tokens
including token $0$. The comparison against ObjectForesight in Table~1 of the
main paper instead follows that method's own convention and drops token $0$.

\subsection{Compute}

Building the training set requires one backbone forward per clip, to the read
step and at stride-2 flow caching; this is a one-off preprocessing cost and is
not part of the inference budget measured in Table~2 of the main paper. Reader
training itself is cheap --- all five folds of one
configuration complete in $16$ minutes on one RTX PRO 6000. All experiments in
the paper and this appendix were run on a single such GPU.

\section{Cross-Validation Protocol}
\label{sec:app:folds}

Every number reported for \METHOD{} is an out-of-fold prediction under
five-fold cross-validation. This section states exactly how the folds are
built, because the \DATASET{} dataset contains related samples that must not be
allowed to straddle a split.

\subsection{Why Grouping Is Necessary}

A single recorded clip contributes more than one training sample. Besides the
recorded clip itself (domain~A), the same first frame and instruction are used
to generate future videos under several sampling seeds, each of which becomes a
domain-B sample supervised by its own generated motion
(Sec.~3 of the main paper). These samples share a first frame, an instruction,
and an object instance. Splitting at the level of individual samples would
therefore place near-duplicates of the same underlying clip on both sides of a
fold boundary, and the resulting scores would overstate generalization.

\subsection{Group Construction}

We assign folds by \emph{source clip}. Every sample, recorded or generated,
carries the identifier of the clip it came from, and that identifier is the
group key, so a recorded clip and every video generated from it collapse to a
single group. Folds are then drawn over \emph{groups}, not samples: the groups are
permuted once under a fixed seed and dealt round-robin into five folds, and
every sample inherits its group's fold.

The resulting split covers $5{,}038$ samples ($2{,}975$ recorded, $2{,}063$
generated) in $2{,}982$ groups, and \textbf{no group spans more than one fold}.
A group holds between one and nine samples: one recorded clip together with
whichever generated variants survived screening. Because groups vary in size,
the folds are not exactly equal in sample count;
Table~\ref{tab:app:folds} gives the realised sizes.

\begin{table}[!t]
\centering
\small
\setlength{\tabcolsep}{1.4mm}
\begin{tabular}{@{}lccccc@{}}
\toprule
Fold & 1 & 2 & 3 & 4 & 5 \\
\midrule
Recorded (A) & $596$ & $597$ & $595$ & $593$ & $594$ \\
Generated (B) & $366$ & $432$ & $355$ & $445$ & $465$ \\
\midrule
Total & $962$ & $1{,}029$ & $950$ & $1{,}038$ & $1{,}059$ \\
\bottomrule
\end{tabular}
\caption{\textbf{Realised fold sizes} under grouping by source clip. Folds are
balanced in recorded samples by construction; the generated counts vary because
a clip contributes as many generated samples as it has surviving generations.}
\label{tab:app:folds}
\end{table}

\subsection{Evaluation}

For each fold we train a Reader on the other four and predict the held-out
fold; the five sets of out-of-fold predictions are then pooled, so every one of
the $5{,}038$ samples is scored exactly once by a model that never saw its
group. Reported errors are per-clip medians over the pooled predictions
(Sec.~\ref{sec:app:metrics}).

Baselines are evaluated on the same fold assignment wherever they are retrained,
so that no comparison is confounded by a difference in the split
(Sec.~\ref{sec:app:baselines}). The exact fold assignment is released with the
code supplement as a single file mapping every sample identifier to its fold
index, so the split can be reproduced without rerunning the grouping code.

\subsection{Filtering Precedes the Split}

Both halves of the dataset are heavily filtered before anything is split
(Sec.~3 of the main paper). On the recorded side, $7{,}246$ pooled candidate
clips are reduced to the $2{,}975$ that carry a meaningful, describable and
gap-free manipulation. On the generated side, $7{,}815$ sampled futures are
labelled by the extraction pipeline and screened
by a human against the recovered track, and $2{,}063$ are retained. Together
they form the $5{,}038$ samples of Table~\ref{tab:app:folds}.

The ordering matters for the validity of the protocol. Screening uses human
judgement, so a screening decision taken \emph{after} the folds were drawn could
in principle be informed by held-out data. Here every filtering decision is made
on the pool as a whole, before any fold exists, and the grouping described above
is then applied to what remains. No filtering step sees a fold boundary, so none
can leak information across one.

\section{Baseline Reproduction}
\label{sec:app:baselines}

Both accuracy baselines are retrained on the \DATASET{} dataset under the fold
assignment of Sec.~\ref{sec:app:folds}. Neither released code ran unmodified on
our data; this section records what had to change, so that the comparison in
Table~1 of the main paper can be audited.

\subsection{ObjectForesight}

ObjectForesight predicts an object pose sequence from multi-frame video
context, the object's CAD mesh, and three ground-truth context poses. We keep
all three inputs --- the comparison is deliberately generous to the baseline on
input, since \METHOD{} sees none of them.

\paragraph{Horizon.}
The method's horizon is tied to its context requirement, and its released HOT3D
configuration predicts $8$ steps at a frame stride of $4$, spanning $1.17$\,s.
Our clips are longer than that, so we run the baseline in two configurations:
its native short horizon, and a longer one ($13$ steps at stride $8$) that
spans the whole action. These are two separately trained models, not two
readings of one model, and each is compared against \METHOD{} evaluated at that
configuration's own timestamps.

\paragraph{Anchor alignment.}
The released data converter anchors each trajectory at a fixed frame offset
within the source clip, whereas our samples are anchored at the start of the
annotated action segment. Comparing the two directly would score the
predictions against a different portion of the motion. We therefore align both
by \emph{absolute} frame index in the source clip rather than by position
within the extracted window. Correcting this alignment changes the measured
gap, and all numbers reported in the main paper are post-correction.

\paragraph{Training.}
Table~\ref{tab:app:of} lists the retraining configuration. The split files are
generated from the same grouped fold assignment used for \METHOD{}, so the two
methods see identical held-out clips.

\begin{table}[!t]
\centering
\small
\setlength{\tabcolsep}{1.2mm}
\begin{tabular}{@{}ll@{}}
\toprule
Setting & Value \\
\midrule
Configuration & released HOT3D config \\
Horizon $H$ / frame stride & $8$ / $4$ \ \ (short) \\
 & $13$ / $8$ \ \ (whole action) \\
Context length & $3$ poses \\
Object library & disabled (mesh supplied per clip) \\
Depth & cached, anchor frame only \\
Batch size & $32$ \\
Learning rate & $1 \times 10^{-4}$ \\
Split & grouped folds (Sec.~\ref{sec:app:folds}) \\
\bottomrule
\end{tabular}
\caption{\textbf{ObjectForesight retraining.} Two horizon configurations are
trained separately; each is evaluated only against \METHOD{} read at its own
timestamps.}
\label{tab:app:of}
\end{table}

\subsection{EgoScaler}

EgoScaler predicts a fixed $20$-step trajectory from a depth map, the object's
initial 6-DoF pose, and a text description. We retrain it
from its released 7B checkpoint on the \DATASET{} dataset. Three issues had to
be resolved before the comparison was meaningful.

\paragraph{Units.}
EgoScaler reports displacement in metres, but its published pipeline recovers
depth from a relative monocular estimator whose output is not metrically
calibrated on egocentric footage; the resulting ``metres'' are inflated by a
roughly constant factor relative to true scale. We therefore do not compare
against its published figure. Instead we re-derive the trajectory in true
metric units on our data, using the dataset's own sensor depth where available,
and report every method in centimetres under one common scale
(Sec.~\ref{sec:app:metrics}).

\paragraph{Normalisation leakage.}
The released evaluation code rescales each predicted trajectory by an amplitude
computed from the \emph{complete ground-truth trajectory} of that clip, which
leaks the motion magnitude the model is supposed to predict. The symptom is
visible in the released protocol's own outputs: final-step error falls below
average error, which cannot happen for an unbiased forecaster whose error grows
with horizon. We remove this rescaling and evaluate the model's raw prediction.

\paragraph{Autoregressive decoding.}
In the released generation path the transformer's cached keys and values are
discarded between decoding steps, so every step after the first is produced
without context. We restore the cache; after the fix all held-out clips decode
to parseable trajectories. Two command-line arguments referenced by the
evaluation entry point are also absent from its parser and were added.

\paragraph{Training.}
We fine-tune the released \mbox{7B} checkpoint end-to-end, unfreezing the
language model, at an effective batch size of $32$ (micro-batch $8$ with
gradient accumulation $4$) for $12$ epochs in bf16 on one GPU, with the
description length capped at $64$ tokens. Target normalisation statistics are
recomputed on our training folds rather than inherited from the released
values, so that training and evaluation are self-consistent.

\subsection{Generate-then-Extract Cascade}

The efficiency baseline in Table~2 of the main paper is a RIGVid-style cascade
that we instantiate ourselves rather than adopt from a released implementation,
so that both routes share a backbone and hardware. It runs the same frozen
Wan2.2-I2V-A14B to a fully denoised and decoded video, then recovers a
trajectory from the generated pixels with an off-the-shelf perception stack:
open-vocabulary detection and segmentation to localise the object, point
tracking through the generated frames, monocular depth to lift the track, and
model-based pose fitting to produce 6-DoF poses. \METHOD{} replaces this entire
route with a single truncated backbone forward and the Reader. Both are timed
on the same GPU over the same clips, and the backbone is configured identically
in both --- same sampler, same guidance, same seed --- so the reported speedup
reflects only the removal of the remaining denoising steps, the VAE decode, and
the perception cascade.

\section{Dataset Statistics}
\label{sec:app:data}

This section expands Sec.~3 of the main paper with the
per-corpus composition and coverage statistics that did not fit there.

\subsection{Composition}

Table~\ref{tab:app:corpus} breaks the $5{,}038$ trajectories down by source
corpus --- HOI4D~\citep{hoi4d}, TACO~\citep{taco}, HOT3D~\citep{hot3d},
OakInk2~\citep{oakink2}, H2O~\citep{h2o} and FPHA~\citep{fpha}.
The recorded half is dominated by HOI4D and HOT3D, the two largest
corpora that supply frame-wise 6-DoF poses for a rigid manipulated object; the
generated half is distributed differently, because a clip contributes generated
samples only in proportion to how many of its sampled futures survive
screening.

\begin{table}[!t]
\centering
\small
\setlength{\tabcolsep}{1.4mm}
\begin{tabular}{@{}lccc@{}}
\toprule
Corpus & Recorded & Generated & Total \\
\midrule
HOI4D    & $1{,}325$ & $1{,}037$ & $2{,}362$ \\
TACO     & $421$ & $508$ & $929$ \\
HOT3D    & $720$ & $7$ & $727$ \\
OakInk2  & $244$ & $226$ & $470$ \\
H2O      & $173$ & $108$ & $281$ \\
FPHA     & $92$ & $177$ & $269$ \\
\midrule
Total & $2{,}975$ & $2{,}063$ & $5{,}038$ \\
\bottomrule
\end{tabular}
\caption{\textbf{\DATASET{} composition by source corpus.} Recorded
trajectories come from the corpus's own 6-DoF annotation; generated ones are
labelled by the extraction pipeline of Sec.~3.3 and screened by
a human.}
\label{tab:app:corpus}
\end{table}

The $2{,}063$ generated trajectories descend from $544$ distinct source clips,
each contributing between one and eight surviving generations: $104$ clips keep
one, $94$ keep two, and the distribution tails off to $18$ clips that keep
eight. This is what makes grouped folds necessary
(Sec.~\ref{sec:app:folds}) --- up to nine samples can share one first frame and
one instruction.

\subsection{Coverage}

\begin{figure*}[!t]
\centering
\includegraphics[width=\linewidth]{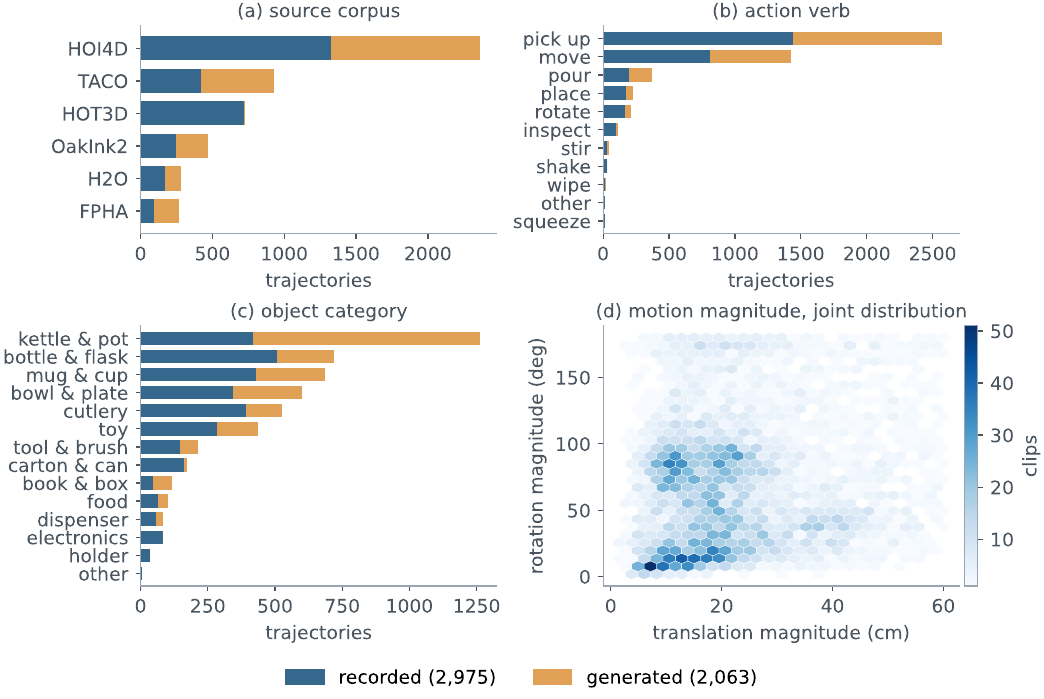}
\caption{\textbf{\DATASET{} statistics.} (a) source corpus, (b) action verb,
(c) object category, each split into recorded and generated trajectories;
(d) joint distribution of per-clip translation and rotation magnitude over all
$5{,}038$ trajectories.}
\label{fig:app:datastats}
\end{figure*}

Fig.~\ref{fig:app:datastats} shows the distribution over source corpora, action
verbs and object categories, together with the joint distribution of
translation and rotation magnitude.

Two properties are worth noting. First, generative augmentation \emph{deepens}
rather than \emph{widens} coverage: every generated trajectory inherits its
source clip's object and instruction, so the corpus retains exactly the
$14$ object categories, $11$ verbs and $92$ object--verb pairs of the recorded
half while packing more trajectories into each cell. Second, panel~(d) shows
that translation and rotation magnitude are only loosely coupled: the corpus
contains both near-pure translations (carrying an object across a table) and
substantial reorientations at small displacement (pouring, inspecting), so a
method cannot score well on both metrics by predicting one from the other.

\subsection{Instruction Annotation}

The source corpora label clips only with coarse action categories, which name
neither which object moves nor how. Every retained clip is therefore
re-annotated from scratch. An AWQ-quantized Qwen3.5-27B drafts an
object-centric label --- the manipulated object, the verb, and the manner of
motion --- from frames sampled across the clip, and a human then corrects that
draft against the video and adjusts the temporal crop so that the clip holds
exactly one complete action. Clips that bundle several actions are split at
this stage rather than discarded.

The resulting instructions are full phrases rather than verb--noun tags, for
example \emph{``pick up the kettle and pour water into the mug on the right''}
in place of \emph{``pour''}. Generated samples inherit the instruction of their
source clip verbatim, which is what makes them a feature-alignment domain
rather than new semantic coverage.

\subsection{Symmetry Handling}

A subset of the manipulated objects are rotationally symmetric about one axis
(bottles, cups, cans), for which the rotation about that axis is unobservable
from RGB and the recorded 6-DoF annotation is arbitrary. Supervising rotation
on those clips would inject noise that no method can fit. We flag these
instances during human screening and, for flagged clips only, project the
target rotation onto the observable subspace before it enters the loss. The
flag is applied identically to the recorded and generated halves, and to every
ablation arm.

\section{Additional Ablations}
\label{sec:app:ablation}

Sec.~5.6 of the main paper reports the ablations that bear directly on the
claims. This section supplies the supporting evidence behind the readout
choices: which attention cell to read, at which denoising step, and which
blocks the scene context is pooled from.

\subsection{The Permutation Control}

Both readout choices are settled with the margin of Eq.~3 in the main
paper rather than with raw correlation, and the control is not a formality.
Manipulation trajectories share a strong temporal profile: an object is picked
up, carried, and set down, so almost any smooth rising-then-falling curve
correlates with almost any ground truth. A readout locked onto that shared
profile alone --- carrying no information about the specific clip --- still
scores $0.43$ against an \emph{unrelated} clip's ground truth. Subtracting the
expected off-diagonal score removes exactly this population prior, so a margin
above zero certifies per-sample signal rather than a shared shape.

\subsection{Head Selection}

We score all $40 \times 40$ (block, head) pairs of the backbone under the
margin. Only three blocks carry a usable object track at all; within them the
ranking is
block $28$ / head $32$ at $0.192$,
block $35$ / head $16$ at $0.184$,
and block $32$ / head $26$ at $0.163$
(head scan, $955$ clips; the same cell scores $0.191$ in the independent step
sweep of Table~3 in the main paper, the difference being run-to-run noise on one
quantity rather than two different quantities).

We confirmed the choice on the end task by retraining the Reader separately on
each of the three candidates: the margin-selected cell gives the lowest ADE. The
training-free criterion and the end-task ranking therefore agree.

\subsection{Readout Step}

\begin{figure*}[!t]
\centering
\includegraphics[width=\linewidth]{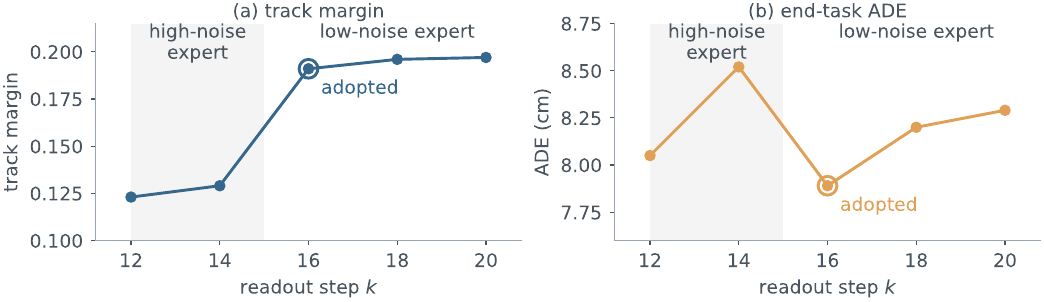}
\caption{\textbf{Readout step sweep.} (a) permutation-controlled track margin
against the recorded 2-D object track; (b) end-task ADE from retraining the
Reader at each step. The shaded region is the backbone's high-noise expert. The
signal appears at the expert switch, not gradually.}
\label{fig:app:readstep}
\end{figure*}

Fig.~\ref{fig:app:readstep} plots the step sweep of Table~3 in the
main paper. The answer is discontinuous rather than gradual: the
margin gains $48\%$ between $k{=}14$ and $k{=}16$ (paired $t = 12.6$ over $955$
clips) and then stops moving, with every later step adding under $3\%$. The
jump lands exactly on the backbone's high- to low-noise expert switch
($t = 900$, between steps $14$ and $16$), which is shaded in the figure. We
adopt $k{=}16$, the earliest step past the switch: reading later costs
additional backbone forwards without recovering more motion signal, and the
end-task ADE in panel~(b) confirms that later steps do not help.

\subsection{Pooled-Hidden Block Selection}

Sec.~5.6 of the main paper establishes that the pooled hidden states matter ---
removing them costs $0.49$\,cm ADE and $0.075$ depth correlation. This section
asks a narrower follow-up question: given three blocks, does it matter
\emph{which} middle block they are read from?

Table~\ref{tab:app:grid} answers it.

\begin{table}[t]
\centering
\small
\setlength{\tabcolsep}{1.1mm}
\begin{tabular}{@{}lcccc@{}}
\toprule
Pooled-hidden blocks & ADE$\downarrow$ & FDE$\downarrow$ & Rot$\downarrow$ & Depth corr.$\uparrow$ \\
\midrule
$\{0, 19, 39\}$ \ (adopted) & $\mathbf{6.26}$ & $\mathbf{9.46}$ & $\mathbf{24.15}$ & $\mathbf{0.660}$ \\
$\{0, 20, 39\}$             & $6.53$ & $9.92$ & $24.39$ & $0.638$ \\
\bottomrule
\end{tabular}
\caption{\textbf{Which middle block to read.} Both rows retrain the Reader with
only the middle block changed, under the grouped five-fold protocol of Table~4
in the main paper. Requiring every block to be available on every sample leaves
$5{,}036$ of the $5{,}038$ samples; the folds are redrawn on that pool, so
absolute values differ slightly from Table~4 while the comparison between the
two rows is exact.}
\label{tab:app:grid}
\end{table}

It does matter. Replacing block $19$ with block $20$ --- one position later in
the network, everything else identical --- degrades all four metrics. The two
blocks are adjacent, so this is not a coarse early-versus-late effect: the
readout is sensitive to where in the stack the scene context is taken from,
which is why the block set is fixed on training folds rather than chosen by
convenience.

\section{Additional Qualitative Results}
\label{sec:app:qual}

\subsection{Extended Comparison}

\begin{figure*}[!t]
\centering
\includegraphics[width=\linewidth]{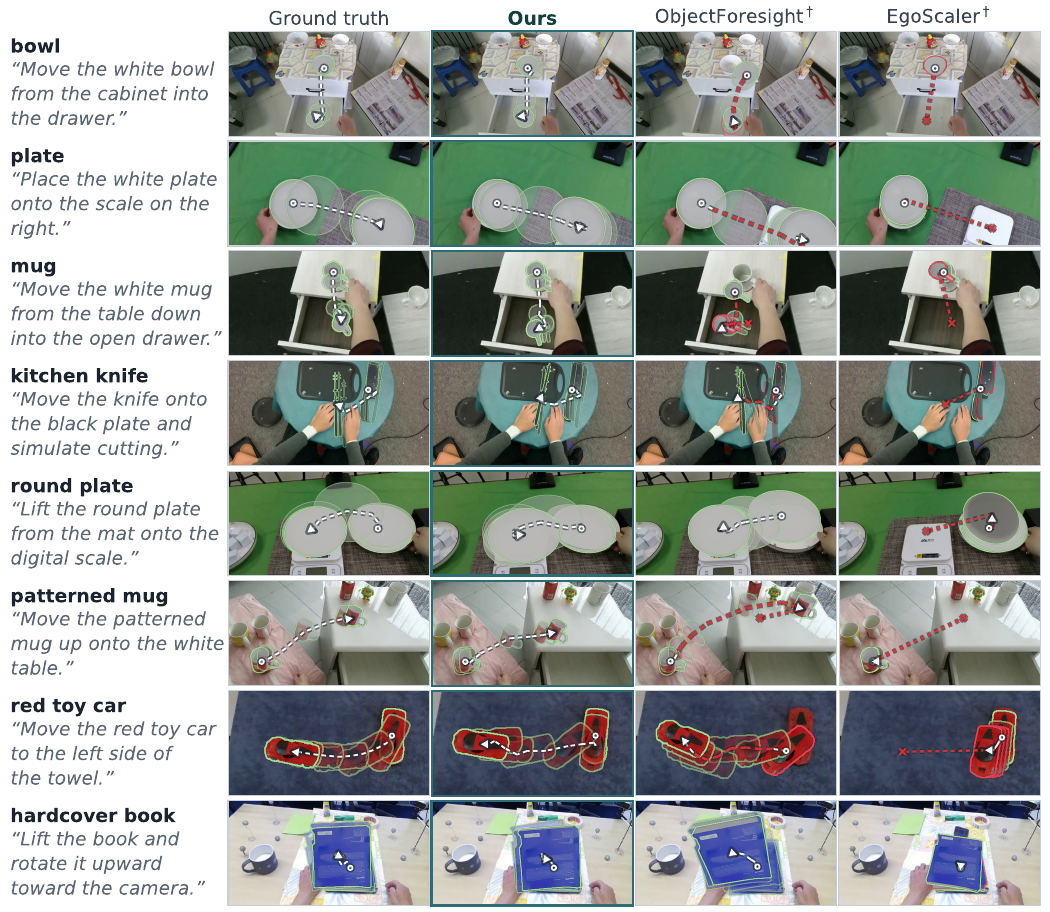}
\caption{\textbf{Extended qualitative comparison.} Each row is one clip, labelled
with the instruction the model was given (abridged for space). Panels overlay $7$
of the $13$ predicted poses on the anchor frame, each outlined in green, with the
dashed line tracing the object's centre. Red marks error: a centroid off by more
than the object's own width, or an orientation off by more than $45^\circ$.
Methods marked $\dagger$ receive privileged input: ObjectForesight is given the
first three poses as ground-truth context, so its overlay starts from the correct
pose by construction. EgoScaler predicts over its native $2$\,s horizon, the
other columns over the whole action.}
\label{fig:app:compare}
\end{figure*}

Fig.~\ref{fig:app:compare} extends the qualitative comparison of the main paper
to more clips, under the identical protocol.

\subsection{Beyond the Training Distribution}

\begin{figure*}[!t]
\centering
\includegraphics[width=\linewidth]{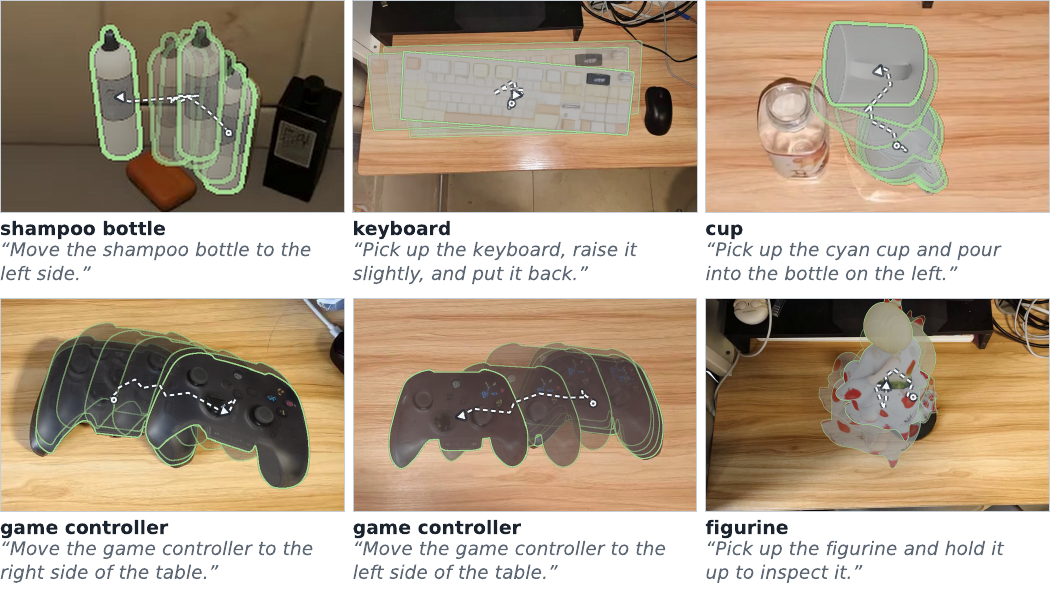}
\caption{\textbf{Generalization beyond the training distribution.} Scenes,
objects and actions absent from every training corpus, rendered as in
Fig.~\ref{fig:app:compare}. The \emph{shampoo bottle} panel is a video-game
frame; the others are phone photographs. Instructions abridged.}
\label{fig:app:wild}
\end{figure*}

Fig.~\ref{fig:app:wild} applies \METHOD{} to further scenes drawn from outside any
of the six source corpora --- phone photographs of everyday desks, and a frame
from a video game. Between them
they cover five distinct object types: a shampoo bottle, a keyboard, a cup, a
game controller and a figurine.
None has a ground-truth trajectory, so these results are qualitative
and are not counted in any quantitative result.

\end{document}